%% file: main.tex
\documentclass[10pt,twocolumn,letterpaper]{article}

\usepackage{cvpr}              

\usepackage{comment}

\input{preamble}

\definecolor{cvprblue}{rgb}{0.21,0.49,0.74}
\usepackage[pagebackref,breaklinks,colorlinks,citecolor=cvprblue]{hyperref}

\def\paperID{8495} 
\def\confName{CVPR}
\def\confYear{2024}

\title{C3Net: Compound Conditioned ControlNet for Multimodal Content Generation}

\author{
    Juntao Zhang\textsuperscript{1*} \quad
    Yuehuai Liu\textsuperscript{1*} \quad
    Yu-Wing Tai\textsuperscript{2} \quad
    Chi-Keung Tang\textsuperscript{1} \\[5pt]
    $^1$HKUST \qquad
    $^2$Dartmouth College \\[8pt]
}

\begin{document}
\twocolumn[{
\renewcommand\twocolumn[1][]{#1}
\maketitle
\begin{center}
    \vspace{-15pt}
    \includegraphics[width=1.0\linewidth]{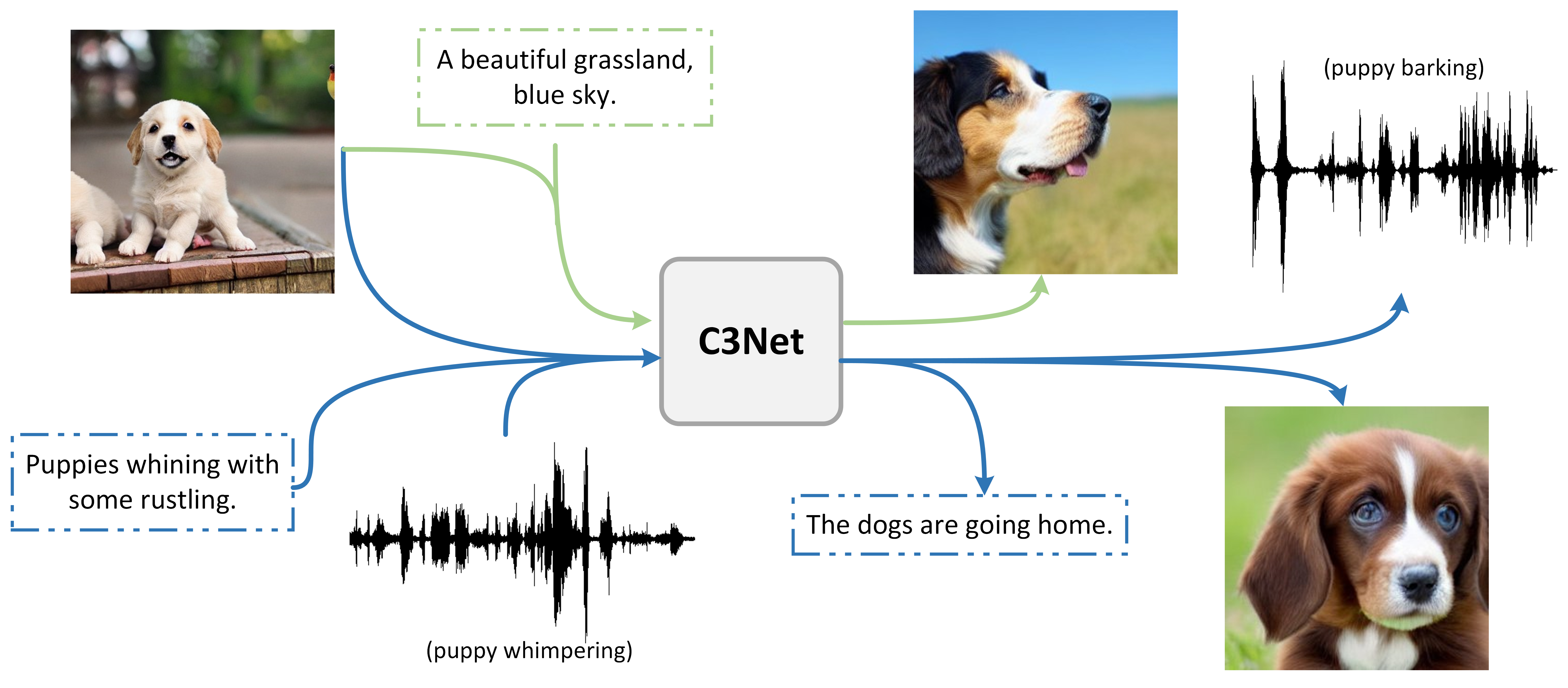}
    \vspace{-17pt}
    \captionsetup{type=figure}
    \caption{{\bf C3Net} generates multimodal contents (e.g., image, audio, and text) taking compound conditions in multiple modalities. The green and blue arrows are two inferences made by C3Net using different combinations of conditions. C3Net can take any combination of image, text, and audio as conditions for content synthesis.}
    \label{fig:teaser}
    \vspace{7pt}
\end{center}
}]

\input{sec/0_abstract}    
\input{sec/1_intro}
\input{sec/2_formatting}
\input{sec/3_finalcopy}
{
    \small
    \bibliographystyle{ieeenat_fullname}
    \bibliography{main}
}
\input{sec/X_suppl}


\end{document}

%% file: preamble.tex
\usepackage[dvipsnames]{xcolor}


%% file: sec/0_abstract.tex
\begin{abstract}
We present Compound Conditioned ControlNet, C3Net, a novel generative neural architecture taking conditions from multiple modalities and synthesizing multimodal contents simultaneously (e.g., image, text, audio). C3Net adapts the ControlNet~\cite{ControlNet} architecture to jointly train and make inferences on a production-ready diffusion model and its trainable copies. Specifically, C3Net first aligns the conditions from multi-modalities to the same semantic latent space using modality-specific encoders based on contrastive training. Then, it generates multimodal outputs based on the aligned latent space, whose semantic information is combined using a ControlNet-like architecture called Control C3-UNet. Correspondingly, with this system design, our model offers an improved solution for joint-modality generation through learning and explaining multimodal conditions, instead of simply taking linear interpolations on the latent space. Meanwhile, as we align conditions to a unified latent space, C3Net only requires one trainable Control C3-UNet to work on multimodal semantic information. Furthermore, our model employs unimodal pretraining on the condition alignment stage, outperforming the non-pretrained alignment even on relatively scarce training data and thus demonstrating high-quality compound condition generation. We contribute the first high-quality tri-modal validation set to validate quantitatively  that C3Net outperforms or is on par with first and contemporary
state-of-the-art 
multimodal generation~\cite{CoDi-ComposableDiffusion}. Our codes  and tri-modal dataset will be released.
\end{abstract}

%% file: sec/1_intro.tex
\section{Introduction}
\label{sec:intro}
Diffusion models have recently emerged as the new state-of-the-art family of deep generative models, with remarkable performance on multimodal modeling~\cite{multimodal_modeling1, CLIP_2, LDM, multimodal_modeling2, multimodal_modeling3}. Correspondingly, we have observed widespread and increasing prevalence of strong cross-modal models that allow generating one single modality from another, including but not limited to text-to-text~\cite{text-to-text1, text-to-text2}, text-to-image~\cite{text-to-image1, text-to-image2, text-to-image3, LDM, text-to-image4} and text-to-audio~\cite{text-to-audio1, text-to-audio2}. However, these existing models cannot simultaneously accept a wider range of input modalities than text or image, nor are they capable of simultaneously generating multiple output modalities in parallel, which leads to limited application in most real-world scenarios where multiple modalities coexist and overlap with one another. The generation capability of each step remains intrinsically constrained even when modality-specific generative models are chained in sequence a multi-step generation setup, which 
can be laborious, time-consuming and compute-demanding. In this regard, 
Composable Diffusion (CoDi)~\cite{CoDi-ComposableDiffusion} is to date the only contemporary work 
capable of concurrently generating any combinations of modalities, simply by taking linear interpolations on the aligned latent space, which results in the downgraded synthesis qualities. Thus, a better and more flexible joint-modality generative model is necessary. 

To achieve better synthesis results while facilitating “any-to-any” generation capabilities, we propose Compound Conditioned ControlNet, or {\em C3Net}, whose
overall architecture design is adapted from 
ControlNet~\cite{ControlNet}, which trains and makes inferences on a production-ready diffusion model and its trainable copies. \textcolor{black}{Our model first aligns the conditions obtained from individual modalities to a shared semantic latent space. During the training of alignment encoders, we utilize unimodal pre-training to mitigate the deficiency of high-quality multimodal datasets.  
The semantic information obtained from individual modalities is further combined through a learnable ControlNet-like architecture called Control C3-UNet. multimodal conditions are then coordinated and merged into the C3-UNet for multimodal synthesis.} Consequently, 
our model can generate multimodal outputs from the aligned latent space.

Thus, {\it 1)} C3Net contributes a better 
solution 
than straightforward linear interpolations on the latent space,  synthesizing more complex and diverse outputs 
beyond them. Notably, C3Net only requires training one Control C3-UNet to work on multimodal conditions, which substantially reduces complexity for joint-modality training and generation. Furthermore, {\it 2)} C3Net employs unimodal pre-training on the condition alignment stage, which facilitates alignment quality even on relatively scarce training data. 
Overall, C3Net outperforms or is on par with the state-of-the-art multimodal generation counterparts, making it the next strong baseline for generating complex and diverse multimodal outputs.


\begin{figure*}[t]
  \centering
    \includegraphics[width=0.98\linewidth]{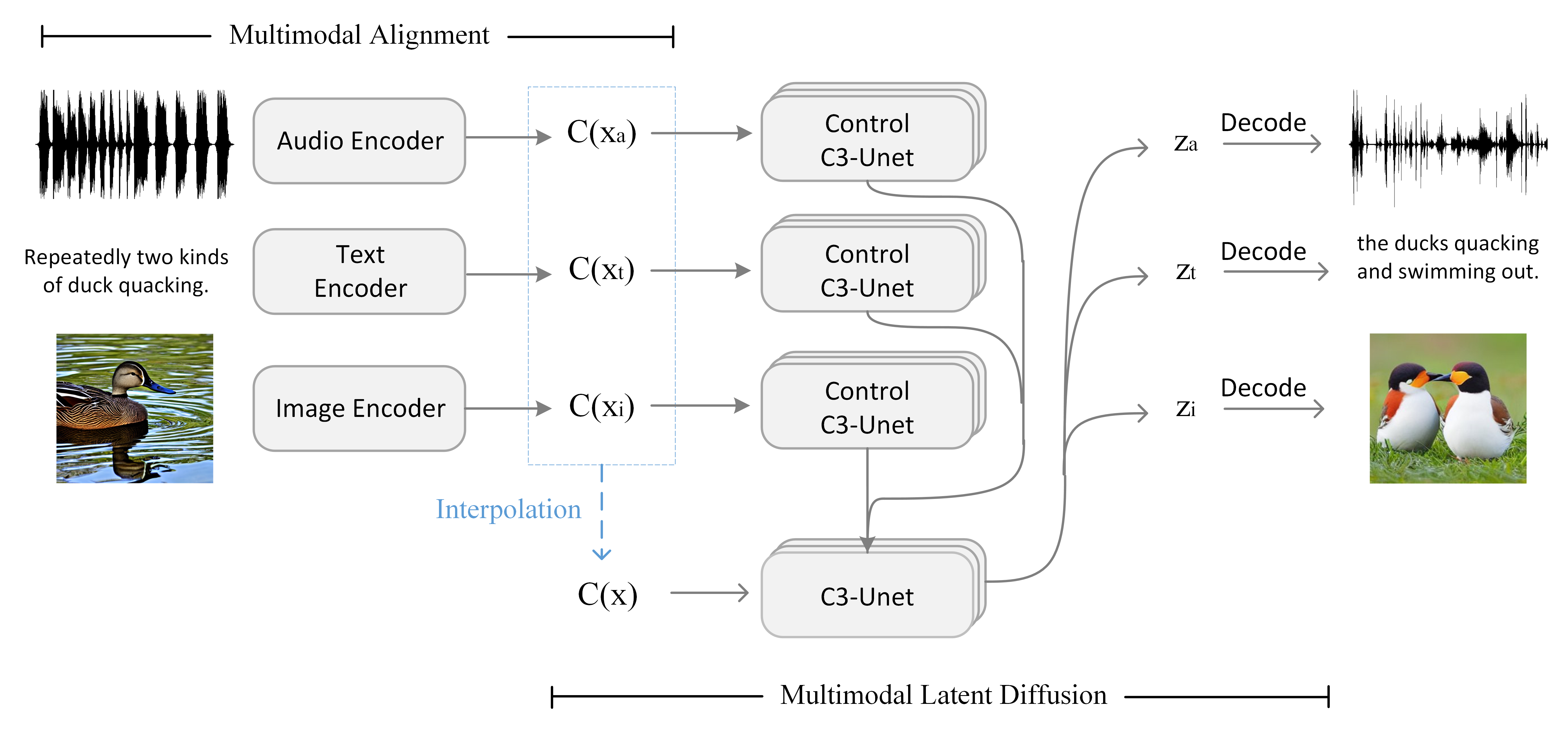}
  \caption{{\bf C3Net} first aligns compound conditions in different modalities to a shared latent space $\xi$, where the encoder takes individuals of compound conditions and generates aligned latent: $C(x_a)$, $C(x_t)$, $C(x_i)$ are all in $\xi$, and $C(x)$ is an interpolation. The aligned condition latent is fed to a generative network consisting of {\em C3-UNet} and {\em Control C3-UNet}, which adaptively learns to coordinate compound conditions in addition to the weighted arithmetic mean in the latent space indicated by the blue dashed box. C3Net generates multiple latent 
  for each modality, denoted as $z_a$, $z_t$, $z_i$, for audio, text, image respectively. Then the $z$'s are decoded using their respective established decoders to generate contents. See Figure~\ref{fig:zoom} for Control C3-UNet and C3-UNet details.}
  \label{fig:overall}
  \vspace{-0.2in}
\end{figure*}

%% file: sec/2_formatting.tex
\section{Related Work}
\label{sec:formatting}

\subsection{Diffusion Models}

\textit{Diffusion models (DMs)} consist of a class of probabilistic generative models capable of understanding the desired data distribution and synthesizing new samples, through a continuous application of denoising autoencoders in output generation. For the three dominant formulations,
\textit{Denoising diffusion probabilistic models (DDPMs)} \cite{DDPM} utilize two Markov Chains for image generation: a forward chain that injects random noise to the data and transforms the data distribution into an unstructured simple prior, and a reverse chain that denoises and recovers the original data by understanding the learnable transition kernels. \textit{Score-based generative models (SGMs)} \cite{SGM, SGMandSDE} introduce score functions defined as the gradient of log probability density, adding a series of escalating Gaussian noise into the data and jointly calculating the scores for all noisy data distributions. \textit{Stochastic Differential Equations (SDEs)} \cite{SGMandSDE} can further be leveraged in the injection and denoising processes, allowing for the scenario of unlimited time steps or noise levels in DDPMs and SGMs. \textit{Latent diffusion models (LDMs)} \cite{LDM} first train a \textit{Variational autoencoder (VAE)} \cite{VAE_1, VAE_2} to encode inputs into a low-dimensional and efficient latent space, and then apply a diffusion model to further generate latent codes. By abstracting negligible details and reducing modeling dimension, the motivation is to focus on the semantic aspects of the data to achieve higher computational efficiency. The diffusion models have achieved state-of-the-art synthesis quality in image inpainting, image superresolution, and audio generation from text. 

\subsection{Composable Diffusion}

\textit{Composable Diffusion (CoDi)} \cite{CoDi-ComposableDiffusion} is a joint-modality generative model capable of producing a combination of output modalities in parallel based on a combination of input modalities, such as text, audio, image and video. CoDi first trains a latent diffusion model for each modality independently, adds a cross-attention module to each diffuser, and further apply an environment encoder to project the latent variables of different LDMs into a shared latent space. CoDi’s design enables multimodal generation without  training on all possible combinations of modalities, reducing the size of training 
from one of exponential to linear.

\subsection{Unimodal Pre-training}

\textit{Unimodal Models} trained on large single-modality datasets can achieve a broader and more diverse coverage of real-world data distribution, without being constrained by the presence of cross-modality data pairs. Specifically, using unimodal models as pre-training can achieve better zero-shot performance compared with the jointly-trained multimodal models, with \textit{MAE} \cite{mae} and \textit{T5} \cite{t5} outperforming the state-of-the-art CLIP-based model under similar model capacities. Moreover, as an effective unimodal pre-training technique for audio processing tasks, \textit{Self-Supervised Audio Spectrogram Transformer (SSAST)} \cite{ssast} enables models to learn the underlying patterns and features of large, unlabeled audio datasets and further improve their performance on the fine-tuning datasets. In the case of C3Net, we first apply unimodal pre-trained encoders for each modality, and then fine-tune the encoders on smaller-scale high-quality datasets based on contrastive learning. 

\subsection{Multimodal Alignment}

\textit{Contrastive Language-Image Pre-Training (CLIP)}~\cite{CLIP_1, CLIP_2} is a neural network that aligns the text and image modalities by pre-training on a large dataset of text-image pairs with a contrastive loss function. Given a sample size of $N$ text-image pairs, CLIP learns to map the two modalities into a common embedding space by jointly training a text encoder and image encoder to maximize the cosine similarity of $N$ matched pairs and minimize the cosine similarity of (\(N^2 - N \)) unmatched pairs using a contrastive loss function. Similar with CLIP, \textit{Contrastive Language-Audio Pre-Training (CLAP)}~\cite{CLAP} aligns the text and audio modalities via contrastive learning paradigm between the audio and text embeddings in pair, also following the same loss function. \textit{CoDi} \cite{CoDi-ComposableDiffusion} proposes the “Bridging Alignment” technique to align conditional encoders for multimodal generation. CoDi leverages CLIP as the pretrained text-image paired encoder, and trains audio and video prompt encoders on audio-text and video-text paired datasets using contrastive learning, with text and image encoder weights frozen. 

The above alignment techniques can be applied to align the latent space of LDMs with multiple modalities to achieve joint multimodal generation. In comparison, C3Net also utilizes the modality-specific encoders to align the conditions from multi-modalities to the same latent space, while it takes a step further by adding neural architecture similar with ControlNet~\cite{ControlNet} to facilitate better understanding of multimodal conditions and joint-modality generation. 

\subsection{ControlNet}

\textit{ControlNet}~\cite{ControlNet} 
learns and adds spatial conditioning to control the large pre-trained diffusion models. By freezing the original weights for the pre-trained diffusion model, ControlNet leverages a trainable copy of its deep-and-robust encoding layers to learn the diverse set of conditional controls and avoid overfitting. The original locked model and the trainable copy are then connected with a zero-initialized convolution layer called “zero-convolution”, where the convolution weights are first initialized to zero and progressively learned throughout  training. 

This architecture provides an effective solution for controlling large diffusion models, while ensuring that no new and harmful noises would be added to the deep features of the diffusion models. In the design of C3Net, we independently apply a ControlNet-like architecture to each input modality, further enabling our model to learn to coordinate multimodal conditions and synthesize more optimized results in the cross-modality generation.

\begin{figure}[ht]
  \centering
    \includegraphics[width=0.99\linewidth]{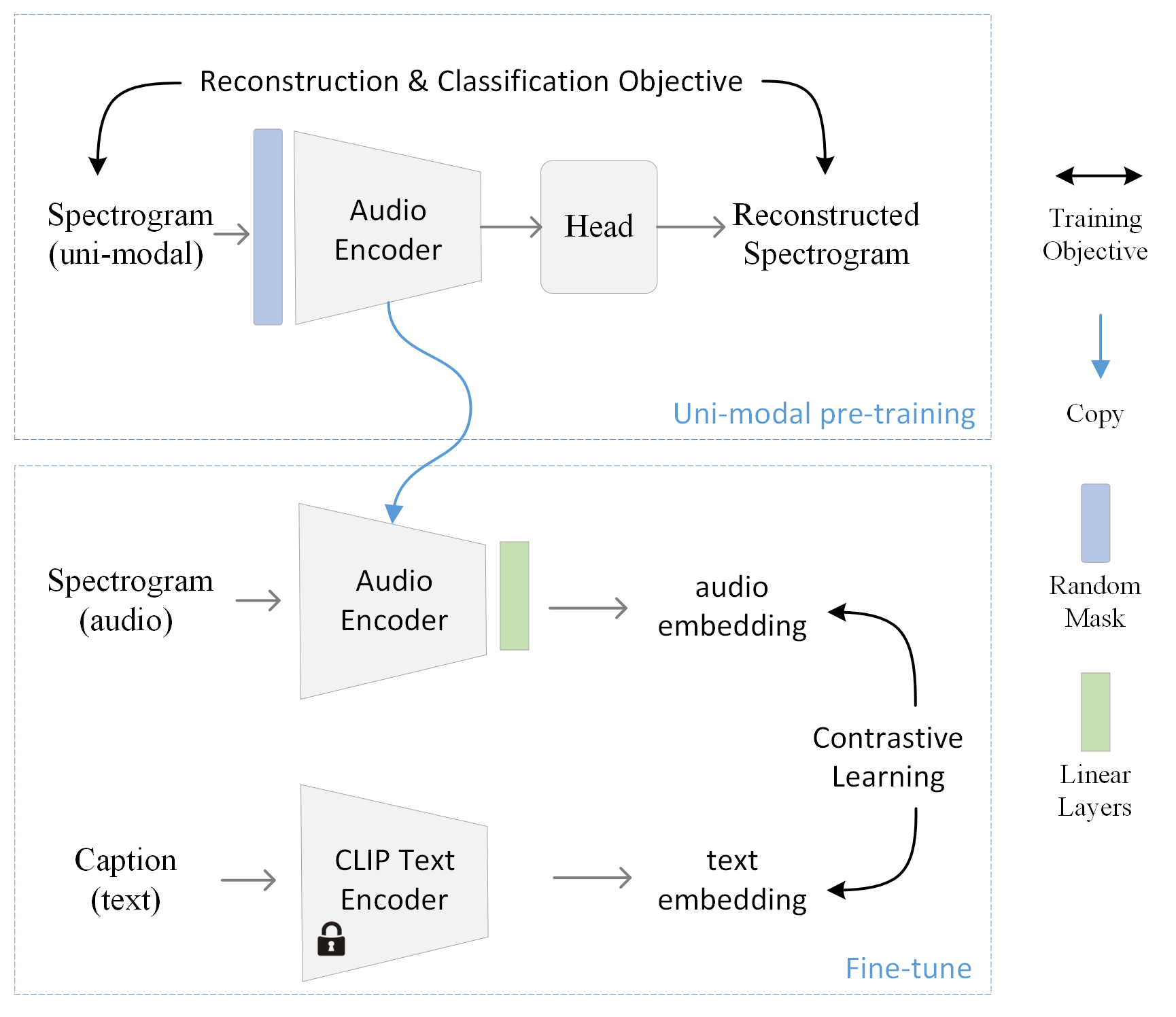}
  \caption{{\bf Uni-modal pre-training} of audio encoder. The audio encoder is first initialized using unsupervised pre-training on large-scale uni-modal data. The encoder is then fine-tuned with objective learning using high-quality multi-modal data.}
  \label{fig:uniPre}
\end{figure}

\section{Method}

C3Net is a neural network architecture for synthesizing multimodal content conditioned on multimodal inputs. Figure~\ref{fig:overall} shows C3Net's overall architecture with content in different modalities (audio, text and image).
We first introduce C3Net's overall structure in Section~\ref{sec:Compound-Conditioned-ControlNet}, including the alignment encoder $C(\cdot)$, and the latent diffusion model consisting of {\em Control C3-Unet} and {\em C3-Unet}. 
Then, we explain the unimodal pre-training for encoders $C(\cdot)$ in Section~\ref{sec:Uni-modal-Pre-trained-Alignment} where, unlike many existing multimodal approaches, \textcolor{black}{multimodal training data (e.g., audio and image pair-up) is not needed in the pre-training stage.}
Finally, we explain {\em C3-UNet} and {\em Control C3-UNet} for compound conditional generation in Section~\ref{sec:Control-C3-UNet}.

\subsection{Compound Conditioned ControlNet}
\label{sec:Compound-Conditioned-ControlNet}
C3Net (Compound Conditioned ControlNet) takes inspiration of the general architecture from~\cite{CoDi-ComposableDiffusion} to enable multimodal generation conditioned on compound information, such as audio, text and image. C3Net first aligns multimodal conditions to a shared latent space. We consider a compound multimodal condition $c_a$, $c_i$, $c_t$, respectively denoting audio, image, and text conditions, and project them to a shared latent space $\xi$ using an encoder $C(\cdot)$. 

\begin{figure}
  \centering
    \includegraphics[width=0.99\linewidth]{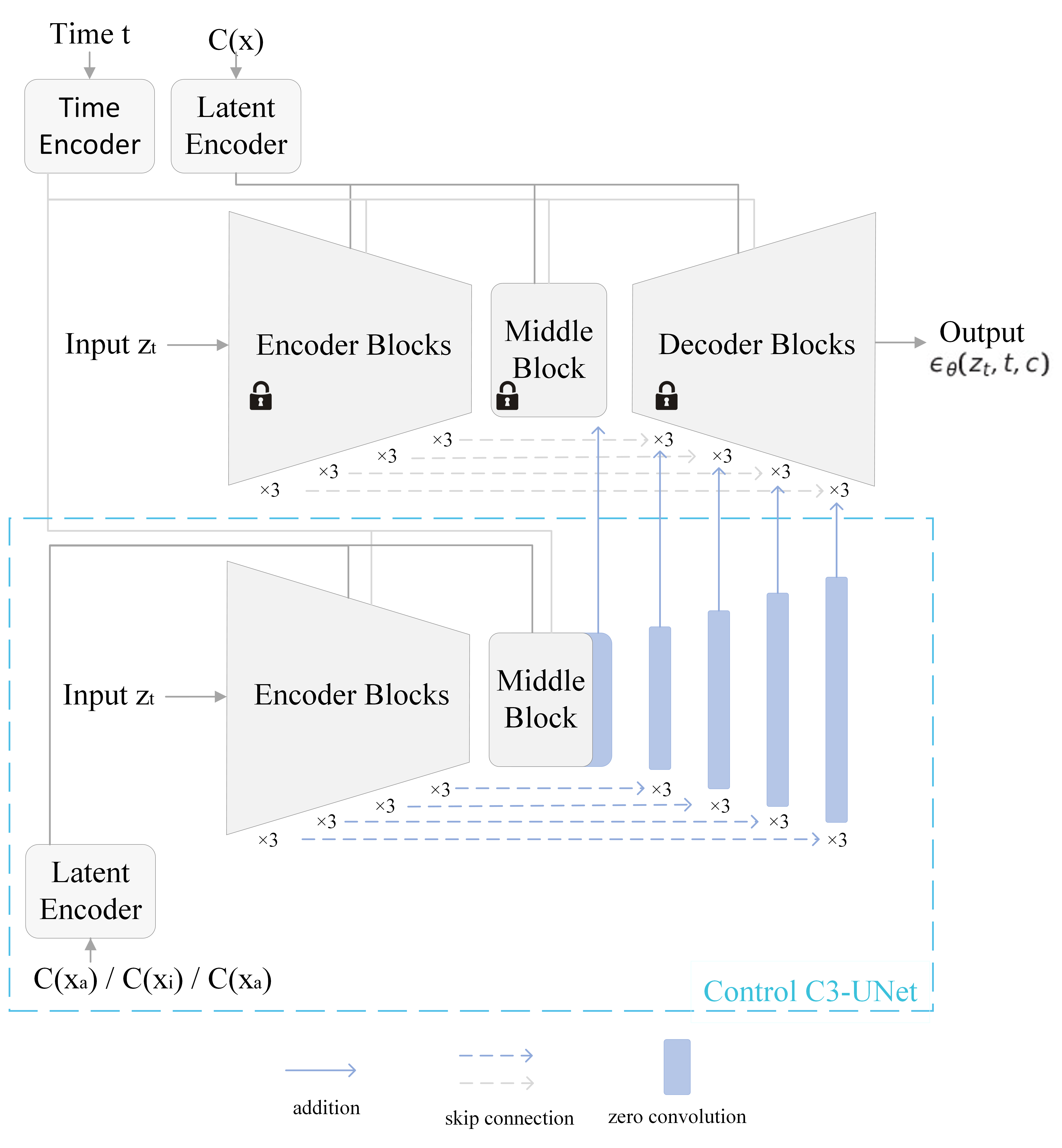}
  \caption{Multimodal generation of  C3Net consists of {\bf C3-UNet} (top) and {\bf Control C3-UNet} (bottom in the blue box). Similar to  ControlNet, Control C3-UNet provides additional control to the C3-UNet. Control C3-UNet takes latent condition aligned from each modality separately and connects to the C3-UNet at each level of skip-connection by addition. $C(x_a)$, $C(x_i)$, $C(x_t)$ are aligned audio, image, text conditions respectively; they are input one at a time. Each modality is generated by its respective UNet pair.}
  \label{fig:zoom}
  \vspace{-0.1in}
\end{figure}

We observed that text captioning is ubiquitously adapted in large-scale multimodal datasets as one of the ground truth labels, and that as mentioned in \cite{CoDi-ComposableDiffusion}, certain dual modalities datasets (e.g., audio and image) are either harshly-aligned or scarce in quantity. To address this issue, choosing a shared latent space $\xi$ in which the text encoder is well-established is advisable. Following the practical implementation of~\cite{CoDi-ComposableDiffusion}, we adopt CLIP~\cite{CLIP_1} latent space as our shared latent space $\xi$. Thus, an instance of the compound condition alignment stage yields a tuple of latent
\begin{equation}
  C(x_a), C(x_i), C(x_t) \in \xi 
\end{equation}
denoting the aligned latent from audio, image, and text conditions, respectively. 

After acquiring latent conditions, we generate multimodal contents using latent diffusion models with C3-UNet and Control C3-UNet as the backbone. Specifically, we can sample feature maps $z_0$ of any modality from a diffusion model sampling process, which is conditioned on $C(x_a), C(x_i), C(x_t)$. Note that the synthesis of different modalities utilizes different diffusion models. Then, $z_0$ is decoded on respective decoders to generate the content of its modality.

\subsection{Uni-modal Pre-trained Alignment}
\label{sec:Uni-modal-Pre-trained-Alignment}

Our encoders $C(\cdot)$ are multi-modal encoders aligning conditions in different modalities to the shared space $\xi$. The encoders are first pre-trained on unimodal datasets and then fine-tuned using contrastive learning proposed in~\cite{CLIP_1}. 
In contrast, the original settings of~\cite{CoDi-ComposableDiffusion} is an alignment model trained from scratch on multi-modal datasets.

We propose to pre-train encoders on {\em unimodal} data because high-quality paired datasets are scarce for some modalities (e.g., audio and text pair). On the other hand, many high-quality unimodal datasets are readily available. In the following, we use the audio encoder as an example. 
As shown in Figure~\ref{fig:uniPre}, we use the pre-trained neural net from~\cite{ssast-unimodal-audio} which is a masked auto-encoder. The MAE has been trained to extract audio features during the unsupervised training stage, which makes it easier for the following contrastive learning for the audio encoder. We then fine-tune the audio encoder using high-quality datasets, which are available on relatively small scales. During the fine-tuning stage of the audio encoder, we utilize an established frozen text encoder from~\cite{CLIP_1}. In detail, we use contrastive objective to fine-tune the pre-trained audio encoder, so that the audio encoder learns to align audio to latent in $\xi$ as similar as possible to the latent that its ground truth caption is aligned to.

Similar to the findings in~\cite{unimodal1, unimodal2, unimodal3, unimodal4, unimodal5, unimodal6}, our encoder networks can primarily learn the data pattern for respective modalities with unsupervised pre-training. Trained with fewer but high-quality multi-modal data, our unimodal pre-trained encoder is on par with or outperformed encoders trained on only paired data on downstream generation tasks.

\subsection{C3-UNet and Control C3-UNet}
\label{sec:Control-C3-UNet}

 Figure~\ref{fig:zoom} shows the multimodal diffusion model of our C3Net, which consists of the C3-UNet and Control C3-UNet.  {\em C3-UNet} is a trained UNet $\mathcal{F}(\cdot,\Theta)$ employed in a latent diffusion model, which generates feature maps conditioned on instances in $\xi$. {\em Control C3-UNet}, similar to the ControlNet setting in \cite{ControlNet}, is a trainable copy $\mathcal{F}(\cdot,\Theta_c)$ of the C3-UNet, where $\Theta_c$ denotes the trainable copy of parameters $\Theta$. In the implementation of C3Net, we use the trained UNet of Composable Diffusion~\cite{CoDi-ComposableDiffusion} as C3-UNet $\mathcal{F}(\cdot,\Theta)$. Figure~\ref{fig:zoom} shows the detailed architecture\footnote{
The trained $\mathcal{F}(\cdot,\Theta)$ is a U-Net with an encoder, a middle block, and a skip-connected decoder. The encoder and decoder contain 12 blocks, and the full model contains 25 blocks, including the middle block. Of the 25 blocks, there are 4 down-sampling and 4 up-sampling blocks. Refer to Figure~\ref{fig:zoom}. 
In C3-UNet, the ``Encoder Blocks" contains 12 encoder blocks in 4 resolutions, while the “×3” indicates the block of the same resolution repeats three times. Condition latent is encoded using the Latent Encoder, and diffusion time steps are encoded with a time encoder using positional encoding. Similar to~\cite{ControlNet}, the Control C3-UNet is a trainable copy of the 12 encoder block and 1 middle block of the C3-UNet. The feature maps are added to the 12 skip-connections and 1 middle block of the C3-UNet after a ``zero convolution" layer.}.

 The Control C3-UNet can provide additional information lost during the latent interpolation. Notably, our Control C3Net takes the aligned latent $C(x_a),C(x_i),$ and $C(x_t)$ separately in each modality. The Control C3-UNet is trained to provide extra information in each condition by modifying the feature maps of the C3-UNet. Thanks to the shared latent space $\xi$, it is sufficient to train {\em one trainable copy of parameters} $\Theta_c$ for the Control C3-UNet. This is because conditions from all modalities have been aligned to the shared $\xi$, and a single set of trained parameters $\Theta_c$ is sufficient for additional control by taking condition latent already aligned in $\xi$ regardless of the original modality.
 
 Our diffusion model follows a similar setting in~\cite{CoDi-ComposableDiffusion} and~\cite{CLIP_1}. The C3-UNet takes a linear interpolation of the aligned latent $C(x_a), C(x_i), C(x_t)$ as the condition. The Control C3-UNet takes conditions in each modality separately and connects to the C3-UNet at each level of skip-connection after multiplying a constant, which we empirically set to be 0.1, but it varies depending on the generation task. Constant multiplied adjusts the additional control scale the Control C3-UNet provides when using different combinations of compound conditions. However, when only a single condition is provided, the Control C3-UNet can not provide additional information and therefore we set the constant to zero.

During training, we use the text-image dataset to train  C3Net's image and text generation, and the text-audio dataset for audio generation (training and validation datasets will be described in Section~\ref{sec:dataset}). Specifically, we train Control C3-UNet on each modality separately. Take image generation as an example, for each image-text pair, denoted as $I$ and $x$ respectively, in the dataset. The ground truth text $x$ is aligned to the shared latent space as $C(x)$, and a masked text $x_m$ is generated by randomly selecting 50\% of the text prompt to replace with empty strings and it is aligned as $C(x_m)$. The C3-UNet takes $C(x_m)$, and Control C3-UNet takes $C(x)$ as condition latent, respectively. The ground truth image $I$ is used to generate $z_0$ in a typical latent diffusion model~\cite{LDM}. We train the C3-UNet to predict noise in a timestep $t$ during the image diffusion. Therefore, the objective function of each modality can be denoted as
\begin{equation}
  \mathcal{L}_c =
    \mathop{\mathbb{E}_{z_0, t, c, \epsilon\in\mathcal{N}(0,1)}}
    \left[ \Vert \epsilon - \epsilon_\theta(z_t, t, c) \Vert ^2_2 \right]
\end{equation}
where $\epsilon$ is the ground truth noise, $\epsilon_\theta(\cdot)$ is the network, and $c$ is the tuple of $C(x_m), C(x)$.

\section{Experiments}
\subsection{Training Datasets}
\label{sec:dataset}
We collected our training datasets for fine-tuning the alignment encoder as well as the respective Control C3-UNet for image, audio, and text synthesis. Major effort was made to clean up flawed data in some datasets through data prepossessing and relation scores, including CLIP score~\cite{CLIPScore-NOT_CLIP}, CLAP~\cite{CLAP} similarity score, and the data quality metrics.

For the fine-tuning of {\em audio encoder}, we used AudioCap~\cite{audiocaps}, a dataset of sounds with event descriptions for audio captioning. We also added the sound effects from Epidemic Sound as provided in~\cite{laion-630k}. We selected 1 million ten-second sound clips from AudioSet~\cite{audioSet} with optimal quality and CLAP similarity score to their text captions.

For the fine-tuning of {\em image and text Control C3-UNets}, we utilized the COCO~\cite{ms-coco} dataset and part of the LAION-400M~\cite{laion-400m} dataset, with both consisting of images and corresponding text captions. For the fine-tuning of {\em audio Control C3-UNet}, we utilized a combination of AudioCap~\cite{audiocaps} and AudioSet~\cite{audioSet} for training.

\subsection{Tri-modality Test Set}
\label{sec:Tri-Modality Test Set}
In the absence of $k$-modality datasets, where $k>2$, for multimodal synthesis evaluation, it is crucial to construct a high-quality evaluation set with three modalities (i.e., image, text, and audio) for evaluating C3Net.

\begin{figure}[h]
  \centering
    \includegraphics[width=0.99\linewidth]{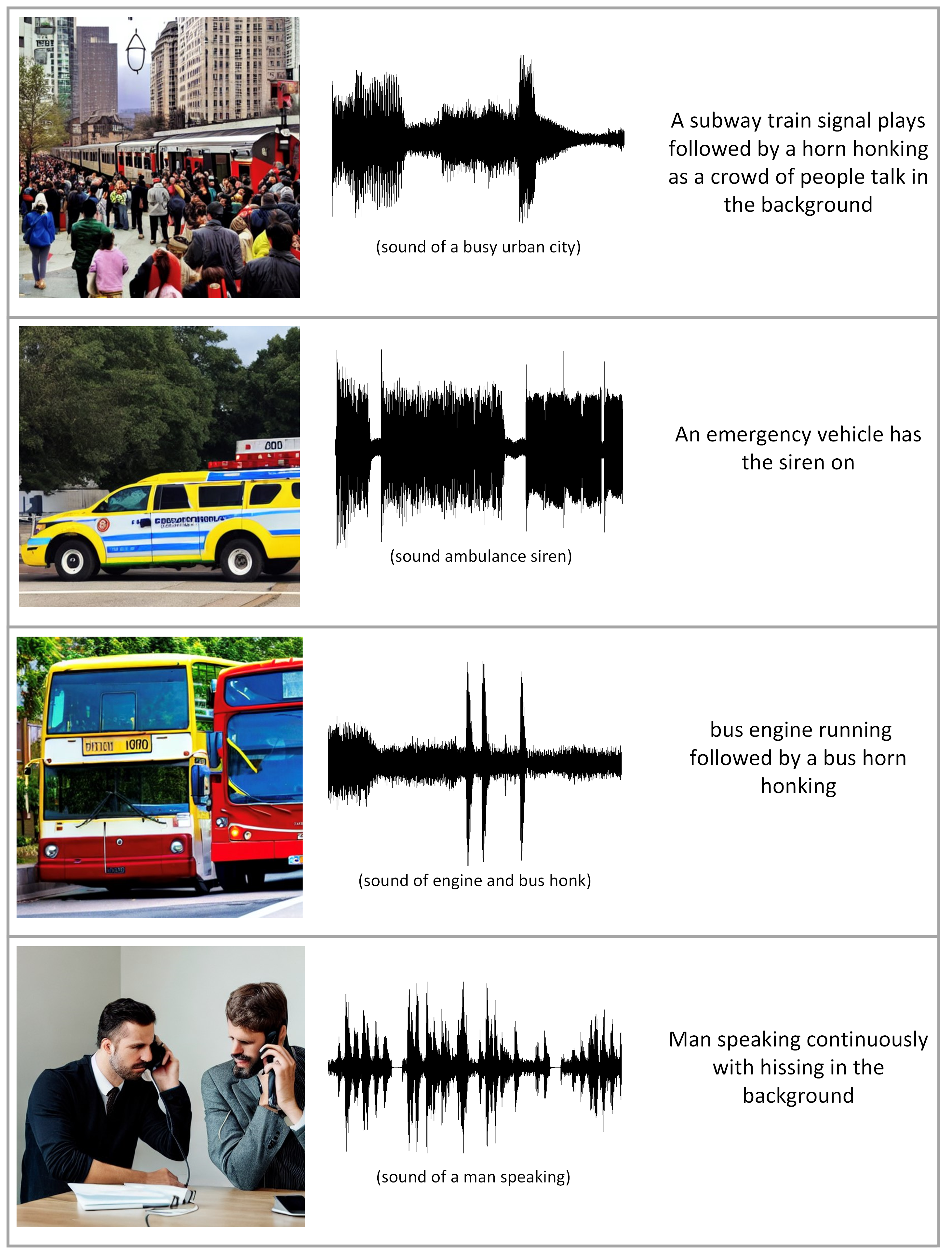}
  \caption{Examples from the Tri-modality Test Set.}
  \label{fig:tri-modality}
\end{figure}

Observing that the AudioCap dataset~\cite{audiocaps} contains high-quality audios and text captions, we curated a tri-modal test set based on AudioCap. Specifically, we first generated the third modality (i.e., image) using Stable Diffusion~\cite{stable-diffusion} prompted on the AudioCap text captions. We further selected 2,000 data tuples based on image quality and CLIP score to ensure that the content for each tuple is highly correlated. As a result, a total of 2,000 tri-modal ground-truth tuples, each with highly relevant audio, image and text caption, are available for evaluating C3Net and CoDi~\cite{CoDi-ComposableDiffusion}, which is to date the most representative (and only) work on diffusion-based tri-modality content generation. We show some examples in Figure~\ref{fig:tri-modality}.

\subsection{Evaluation Results}
\begin{figure*}[t]
  \centering
    \includegraphics[width=0.95\linewidth]{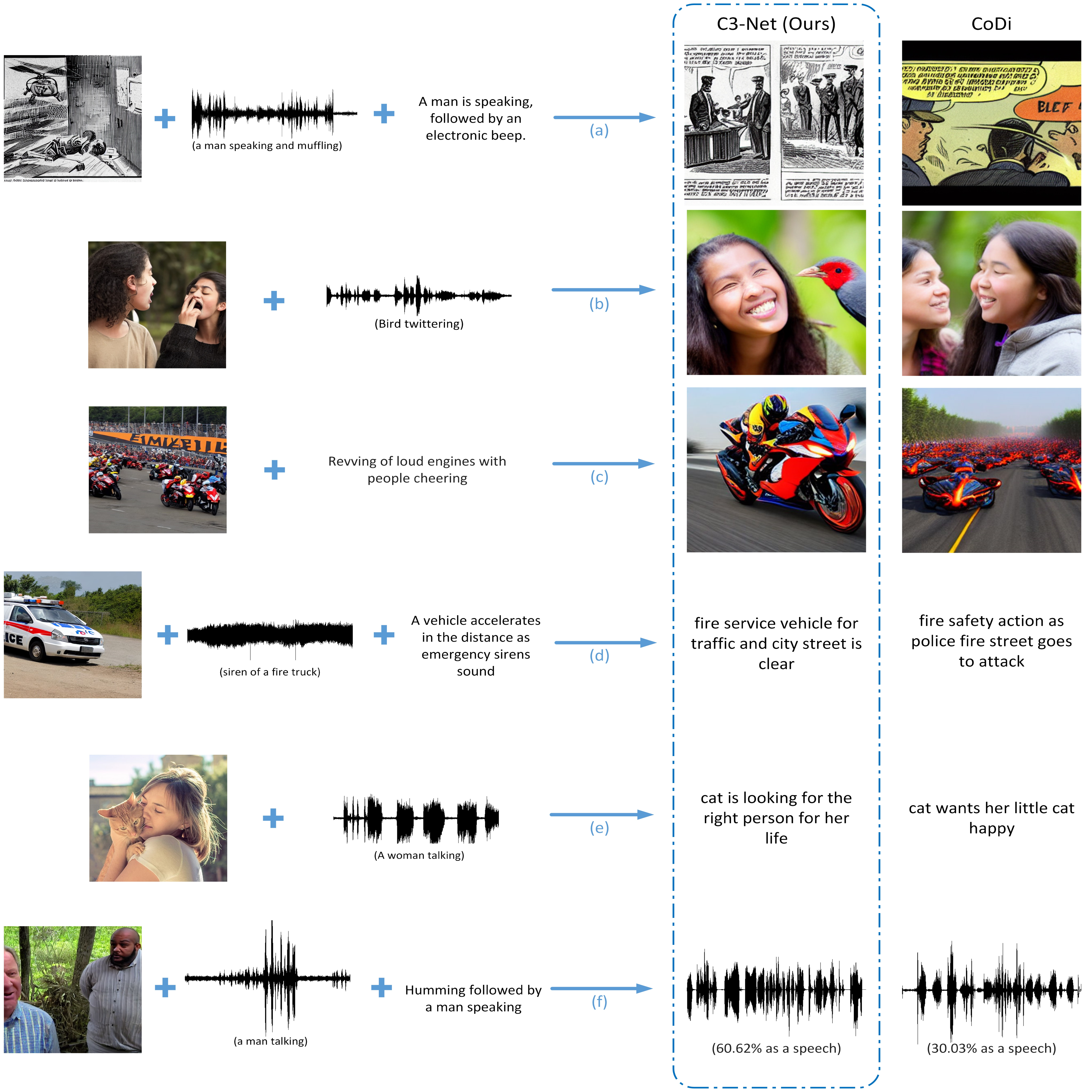}
 
  \caption{{\bf Qualitative comparison} of compound-conditioned synthesis. The examples are conditioned on two or more images, texts, and audio conditions. {\it (a)} C3Net optimally extracts the feature of the blank-and-white sketch in the image condition. {\it (b)} C3Net better utilizes the audio condition, the sound of birds twittering. {\it (c)} C3Net generates an image of higher quality by focusing on the main subjects in the text condition. {\it (d)} The synthesized caption form C3Net has subject {\em fire service vehicle}, which is the optimal combination of subjects in all conditions. {\it (e)} C3Net synthesizes text including {\em person} and {\em cat}, where the baseline generation only has {\em cat}. This may be because the simple interpolation used in CoDi mixes the cat and person features into one subject. {\it (f)} The synthesized audio from C3Net rates higher in probability as a piece of speech classified by the model from~\cite{beats}.}
  \label{fig:compare}
  \vspace{-0.2in}
\end{figure*}

Figure~\ref{fig:compare} shows qualitative comparison on compound condition image synthesis between C3Net and our baseline. 
We will show quantitative comparison in the following.

\subsubsection{Unimodal Pre-training}
We evaluate the unimodal pre-training results by comparing the image and text generated by C3Net and CoDi~\cite{CoDi-ComposableDiffusion}. Evaluation is conducted on the AudioCap~\cite{audiocaps} test set, with which we generate text captions and images conditioned on the ground truth audio. Table~\ref{tab:uni-text} shows the correlation between the generated text and its ground truth captions, assessed by CIDEr-D~\cite{cider-d} and SPIDEr~\cite{SPIDEr}. Table~\ref{tab:uni-image} tabulates the image synthesis quality assessed by the Inception Score, as well as the correlation between the generated images and ground truth text captions assessed by the CLIP score~\cite{CLIPScore-NOT_CLIP}. Evaluations on audio synthesis are not available in this case, as the model design of CoDi does not support audio generation when taking audio as a condition.
Note that in the scenario of taking an audio as the only condition, C3Net and CoDi differ only in the alignment stage, which makes it an ideal ablation study. Under such settings, C3Net applies an audio encoder pre-trained on unimodal data, while CoDi uses an audio encoder without unimodal pre-training. 
\begin{table}[h]
\small
  \centering
  \begin{tabular}{c c c}
    \toprule
    Method & CIDEr-D $\uparrow$ & SPIDEr $\uparrow$\\
    \midrule
    CoDi & 0.0654 & 0.0608 \\
    C3Net (Ours) & {\bf 0.0704} & {\bf 0.0622} \\
    \bottomrule
  \end{tabular}
  \caption{Unimodal pre-training assessed by the correlation between the synthesized {\bf texts} and ground truth text captions on the AudioCap test set. Comparisons are made between  C3Net (with unimodal pre-trained encoders) and CoDi (without as such).}
  \label{tab:uni-text}
  \vspace{-0.2in}
\end{table}
\begin{table}[h]
\small
  \centering
  \begin{tabular}{c c c}
    \toprule
    Method & Inception Score $\uparrow$ & CLIP $\uparrow$\\
    \midrule
    CoDi & 1.7730, \;0.1450 & 23.192\\
    C3Net (Ours) & {\bf 1.7732}, \;{\bf 0.1535} & {\bf 23.325}\\
    \bottomrule
  \end{tabular}
  \caption{Unimodal pre-training assessed by Inception Score of the generated {\bf images}, and the CLIP score between the generated {\bf images} and the ground truth text captions. Images are synthesized conditioning on the AudioSet test set audio.}
  \label{tab:uni-image}
  \vspace{-0.2in}
\end{table}

\subsubsection{Multimodal Synthesis}
We evaluate the multimodal synthesis capabilities of C3Net on the Tri-modality Test Set introduced in~\ref{sec:Tri-Modality Test Set}. 
To assess the synthesis quality on compound conditions, we generate images, text, and audio conditioned on each respective tuple within the Tri-modality Test Set. 

To evaluate {\bf image} synthesis, we measure the Fréchet inception distance~\cite{FID-Frechet_inception_distance} between the synthesized image and its ground truth image as well as the CLIP score~\cite{CLIPScore-NOT_CLIP} between the generated image and its ground truth text caption. Table~\ref{tab:tri-image} shows that C3Net generates images that relate closer to both the text and image conditions, demonstrating that our Control C3-UNet architecture offers a more optimized solution for compound condition image synthesis.
\begin{table}[h]
\small
  \centering
  \begin{tabular}{c c c c c c c}
    \toprule
    Method & FID $\downarrow$ & CLIP $\uparrow$\\
    \midrule
    CoDi & 11.39 & 25.17\\
    C3Net (Ours) & {\bf 10.97} & {\bf 25.29}\\
    \bottomrule
  \end{tabular}
  
  \caption{Compound conditioned {\bf image} synthesis assessed on the Tri-modality Test Set. Generation quality is measured by the Fréchet inception distance between the synthesized image and its ground truth image, and the CLIP score between the synthesized image and its ground truth text caption.}
  \label{tab:tri-image}
\end{table}

To evaluate {\bf text} synthesis, we measure the caption correlation metrics between the synthesized text and the ground truth captions. The caption metrics include BLEU-1~\cite{bleu}, ROUGE-L~\cite{rouge-l}, CIDEr-D~\cite{cider-d}, and SPIDEr~\cite{SPIDEr}. Table~\ref{tab:tri-text} shows that C3Net generates text outputs more closely correlated with the ground truth compared to CoDi.

\begin{table}[h]
  \centering
  \resizebox{\columnwidth}{!}{
  \begin{tabular}{c c c c c c c}
    \toprule
    Method &  BLEU-1 $\uparrow$ & ROUGE-L $\uparrow$ & CIDEr-D  $\uparrow$ & SPIDEr  $\uparrow$\\
    \midrule
    CoDi & 0.1059 & 0.1019 & 0.0651 & 0.0631 \\
    C3Net (Ours) & {\bf 0.1104} & {\bf 0.1045} & {\bf 0.0713} & {\bf 0.0665} \\
    \bottomrule
  \end{tabular}
  }
  \caption{{\bf Text} synthesis assessed on the Tri-modality Test Set. We evaluate the correlation between synthesized texts and ground truth text captions using a variety of caption metrics.}
  \label{tab:tri-text}
  \vspace{-0.1in}
\end{table}

To evaluate {\bf audio} synthesis, we measure OVL (Overall Impression), REL (Text Relevant) similar with the settings in~\cite{AudioGen-subjective}, and FAD~\cite{FAD-frechet_audio} to evaluate the audio quality. As shown in Table~\ref{tab:tri-audio}, C3Net outperforms CoDi in terms of OVL and REL. When measuring the correlation between the synthesized audio and the ground truth audio, C3Net yields a slightly weaker FAD score compared to CoDi. 

\begin{table}[h]
\small
  \centering
  \begin{tabular}{c c c c}
    \toprule
    Method & OVL $\uparrow$& REL $\uparrow$& FAD $\downarrow$\\
    \midrule
    Reference & 81.07 & 79.31 & -\\
    CoDi & 62.91 & 59.01 & 11.4\\
    C3Net (Ours) & {\bf 63.25} & {\bf 59.83} & 11.7\\
    \bottomrule
  \end{tabular}
  \caption{{\bf Audio} synthesis assessed on the Tri-modality Test Set. The audio evaluation metrics include OVL and REL between the synthesized audios and the ground truth captions. We also evaluated the FAD between generated audio and its ground truth.
  }
  \label{tab:tri-audio}
  \vspace{-0.2in}
\end{table}

\subsubsection{Synthesized Audio Classification}
To further assess the quality of audio synthesized by C3Net, we compare the classification accuracy of the generated image conditioned on the audio-text pairs in the ESC-50~\cite{ESC-50} dataset. In this experiment, we first synthesized audio conditioned on the audio and text pairs. Then, we classified the generated audio using the classification model given in~\cite{beats}. Table~\ref{tab:audio_class} tabulates the results, where a higher accuracy indicates more optimized audio synthesis on compound conditions, which keeps the shared features in multimodal conditions.
\begin{table}[h]
\small
  \centering

  \begin{tabular}{@{}lcc@{}}
    \toprule
    &  Codi & C3Net (Ours) \\
    \midrule
    Accuracy (\%) & 21.05 & {\bf 23.25} \\
    \bottomrule
  \end{tabular}
  \caption{Classification accuracy on synthesized {\bf audio} conditioned on audio-text pairs in ESC-50. A higher accuracy indicates a better ability to keep shared features in multimodal conditions.}
  \label{tab:audio_class}  
  \vspace{-0.2in}
\end{table}

%% file: sec/3_finalcopy.tex
\section{Conclusion and Discussion}
In this paper, we propose C3Net, a multimodal generative model conditioned on compound content, which applies unsupervised pre-training on unimodal datasets and further leverages a ControlNet-like architecture to coordinate compound conditions. Through extensive experiments, we demonstrate that C3Net is capable of synthesizing high-quality multimodal contents on compound conditions by coordinating them through a learnable process, and addressing the deficiencies of datasets through unimodal pre-training.

While C3Net has shown remarkable progress in joint-modality generation, there exist remaining challenges that need to be addressed in the future. One of the issues is the choice of the shared latent space, such as the CLIP~\cite{CLIP_1} latent, which may not be optimal for all modalities, particularly audio. 
To address this issue, a contrastive learning process that takes into account multiple modalities may be more effective. Another challenge is that aligning latent conditions using contrastive learning may sacrifice the unique information contained in a modality, as noted in a previous study~\cite{in-cross-Modality}. One solution to this issue is to use a similar alignment objective as proposed in~\cite{in-cross-Modality}, which aims to construct more meaningful latent modality structures. Addressing these challenges can improve the effectiveness of multimodal generative models, leading to more advanced and sophisticated content synthesis in the future.



%% file: sec/X_suppl.tex
\clearpage
\setcounter{page}{1}

\twocolumn[{
\renewcommand\twocolumn[1][]{#1}
\maketitlesupplementary
\begin{center}
    \vspace{-15pt}
    \includegraphics[width=1.0\linewidth]{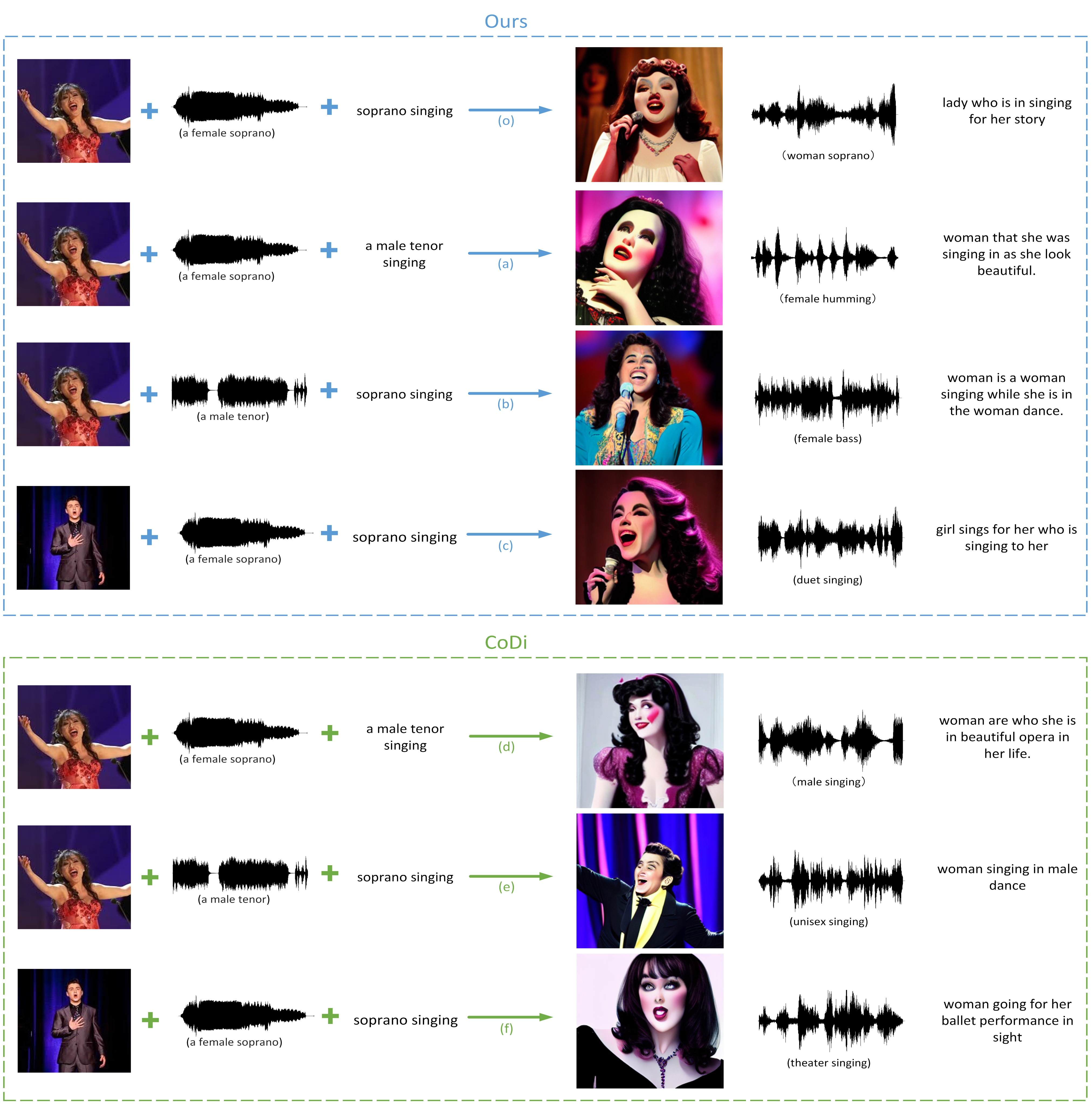}
    \vspace{-17pt}
    \captionsetup{type=figure}
    \caption{Further experiments in the following settings are conducted to investigate the effect of multimodal conditions which are in conflict or contradictory among each other. In {\it (a)}, the text condition relates to a male tenor, while the image and audio depict a female soprano. In {\it (b)}, male tenor audio is used as a condition, while the image and text respectively relate to a female soprano. In {\it (c)}, the image indicates a male singer, while the audio and text describe a female soprano. In {\it (o)}, which is a control, we use consistent conditions to compare the differences in the above experimental scenarios. We repeated the  experiment with same settings and produce  generations using the baseline~\cite{CoDi-ComposableDiffusion}. }
    \label{fig:sup}
    \vspace{7pt}
\end{center}
}]

\section{Robustness to Contradictory Multimodal Conditions}

We further explored how contradictory conditions can influence the generation of C3Net. Through extensive experiments, we found that rather than causing collapse and generating nonsense or dominated by a single condition, C3Net coordinates contradictory inputs innovatively. 
We elaborate on C3Net’s robustness by an example in Figure~\ref{fig:sup}, where all other conditions indicate a female soprano except one, which describes a male tenor.

The example shows four scenarios, including one control and three experimental scenarios, where 
two conditions describe a female soprano, while the remaining one relates to a male tensor. C3Net generates images, audio, and text conditioned on these contradictory inputs. The generated images and text all describe a female, the most frequent subject in three conditions. In scenarios {\it (a)} and {\it (b)}, the discrepancy changes the tone of generated audio from a female soprano to a lower frequency humming. In {\it (c)}, the generated audio and text indicate another singer, coordinating the condition of {\em female soprano} and {\em male tenor}. In {\it (o)}, a control, we use consistent conditions to compare the differences in the above experimental scenarios.

We repeated the  experiment with the same settings to produce multi-modal generations using the baseline~\cite{CoDi-ComposableDiffusion}. We found that without the Control C3-UNet structure, the generated contents tend to be intermediates of the contradictory conditions (e.g., scenario {\it (e)}) or shifted in meaning (e.g., text generated in scenario {\it (a)} and {\it (e)}). These defects are likely resulted from using simple interpolation to coordinate multiple conditions. 

\section{Audio Files}
We put in our supplemental zip file all the relevant audio files depicted in each figure in the main paper and this supplementary document, including all condition audios and generated audios.